\documentclass[11pt,a4paper,table]{article}
\usepackage[hyperref]{acl2019}
\usepackage{times}
\usepackage{latexsym}
\usepackage{microtype}

\usepackage[T5,T2A,T1]{fontenc}
\usepackage[utf8]{inputenc}

\usepackage[T1]{fontenc}    
\usepackage[utf8]{inputenc} 
\usepackage{caption}

\usepackage{amsfonts}       
\usepackage{amsmath}
\usepackage{array}
\usepackage{arydshln}
\usepackage{booktabs}       
\usepackage{calc}
\usepackage{enumitem}
\usepackage{graphicx}
\usepackage{hyperref}       
\usepackage{lipsum}
\usepackage{microtype}      
\usepackage{nicefrac}       
\usepackage{stfloats}
\usepackage{tabularx}
\usepackage{url}            
\usepackage{xcolor}
\usepackage{subcaption}

\aclfinalcopy

\renewcommand*{\arraystretch}{1.5}

\ifaclfinal
    \newcommand\encoder{polyai-encoder}
\else
    \newcommand\encoder{encoder}
\fi

\title{
    A Repository of Conversational Datasets \\
\ifaclfinal
    {
        \ttfamily \small \href{https://github.com/PolyAI-LDN/conversational-datasets}{github.com/PolyAI-LDN/conversational-datasets}
    }
\else 
	{
        \ttfamily \small \href{https://github.com/AAAA/BBBB}{github.com/AAAA/BBBB}
    }
\fi
}

 \author{
 Matthew Henderson,
 Pawe{\l} Budzianowski,
 I{\~{n}}igo Casanueva,
 Sam Coope,
 Daniela Gerz, \\
 {\bf Girish Kumar,
 Nikola Mrk{\v{s}}i\'c,
 Georgios Spithourakis,
 Pei-Hao Su,
 Ivan Vuli\'{c}, and
 Tsung-Hsien Wen
 } \\
 \texttt{\small matt@poly-ai.com} \\
 PolyAI Limited,
 London, UK.
}

\begin{document}

\maketitle

\begin{abstract}
    Progress in Machine Learning is often driven by the availability of large datasets, and consistent evaluation metrics for comparing modeling approaches. To this end, we present a repository of conversational datasets consisting of hundreds of millions of examples, and a standardised evaluation procedure for conversational response selection models using \emph{1-of-100 accuracy}. The repository contains scripts that allow researchers to reproduce the standard datasets, or to adapt the pre-processing and data filtering steps to their needs. We introduce and evaluate several competitive baselines for conversational response selection, whose implementations are shared in the repository, as well as a neural encoder model that is trained on the entire training set.
\end{abstract}

\section{Introduction}

Dialogue systems, sometimes referred to as conversational systems or conversational agents, are useful in a wide array of applications. They are used to assist users in accomplishing well-defined tasks such as finding and/or booking flights and restaurants \cite{Hemphill:1990,Williams:12,ElAsri:2017sigdial}, or to provide tourist information \cite{Henderson:14b,Budzianowski:2018emnlp}. They have found applications in entertainment \cite{Fraser:2018iva}, language learning \cite{Raux:2003,Chen:2017survey}, and healthcare \cite{Laranjo:2018,Fadhil:2019arxiv}. Conversational systems can also be used to aid in customer service\footnote{For an overview, see \ttfamily \scriptsize \href{https://poly-ai.com/blog/towards-ai-assisted-customer-support-automation}{poly-ai.com/blog/towards-ai-{\allowbreak}assisted-customer-support-automation}}
or to provide the foundation for intelligent virtual assistants such as Amazon Alexa, Google Assistant, or Apple Siri.

Modern approaches to constructing dialogue systems are almost exclusively data-driven, supported by modular or end-to-end machine learning frameworks \cite[\textit{inter alia}]{young:10b,VinyalsL15,Wen:2015emnlp,Wen:2017icml,Wen:17,Mrksic:2018acl,Ramadan:2018acl,Li:2018arxiv}. The research community, as in any machine learning field, benefits from large datasets and standardised evaluation metrics for tracking and comparing different models.
However, collecting data to train data-driven dialogue systems has proven notoriously difficult. First, system designers must construct an ontology to define the constrained set of actions and conversations that the system can support \cite{Henderson:14a,Henderson:14b,Mrksic:15}. Furthermore, task-oriented dialogue data must be labeled with highly domain-specific dialogue annotations \cite{ElAsri:2017sigdial,Budzianowski:2018emnlp}. Because of this, such annotated dialogue datasets remain scarce, and limited in both their size and in the number of domains they cover. For instance, the recently published MultiWOZ dataset \cite{Budzianowski:2018emnlp} contains a total of 115,424 dialogue turns scattered over 7 target domains. Other standard task-based datasets are typically single-domain and smaller by several orders of magnitude: DSTC2 \cite{Henderson2014a} contains 23,354 turns, Frames \cite{ElAsri:2017sigdial} comprises 19,986 turns, and M2M \cite{Shah:2018naacl} spans 14,796 turns.

An alternative solution is to leverage larger conversational datasets available online. Such datasets provide natural conversational structure, that is, the inherent context-to-response relationship which is vital for dialogue modeling. In this work, we present a \textit{public} repository of three large and diverse conversational datasets containing hundreds of millions of conversation examples. Compared to the most popular conversational datasets used in prior work, such as length-restricted Twitter conversations \cite{Ritter:2010naacl} or very technical domain-restricted technical chats from the Ubuntu corpus \cite{Lowe2015TheUD,Lowe:2017dd,gunasekara2019dstc7}, conversations from the three conversational datasets available in the repository are more natural and diverse. What is more, the datasets are large: for instance, after preprocessing around 3.7B comments from Reddit available in 256M conversational threads, we obtain 727M valid context-response pairs. Similarly, the number of valid pairs in the OpenSubtitles dataset is 316 million. To put these numbers into perspective, the frequently used Ubuntu corpus v2.0 comprises around 4M dialogue turns. Furthermore, our Reddit corpus includes 2 more years of data and so is substantially larger than the previous Reddit dataset of \newcite{AlRfou:2016arxiv}, which spans around 2.1B comments and 133M conversational threads, and is not publicly available.

Besides the repository of large datasets, another key contribution of this work is the common evaluation framework. We propose applying consistent data filtering and preprocessing to public datasets, and a simple evaluation metric for response selection, which will facilitate direct comparisons between models from different research groups.

These large conversational datasets may support modeling across a large spectrum of natural conversational domains. Similar to the recent work on language model pretraining for diverse NLP applications \cite{Howard:2018acl,Devlin:2018arxiv,Lample:2019arxiv}, we believe that these datasets can be used in future work to pretrain large general-domain conversational models that are then fine-tuned towards specific tasks using much smaller amounts of task-specific conversational data. We hope that the presented repository, containing a set of strong baseline models and standardised modes of evaluation, will provide means and guidance to the development of next-generation conversational systems.

\ifaclfinal
The repository is available at {
    \ttfamily \small \href{https://github.com/PolyAI-LDN/conversational-datasets}{github.com/{\allowbreak}PolyAI-LDN/conversational-datasets}.
}
\else
The repository is available at {
    \ttfamily \small \href{https://github.com/AAAA/BBBB}{github.com/{\allowbreak}AAAA/BBBB}.
}
\fi
\label{s:introduction}

\section{Conversational Dataset Format}

Datasets are stored as Tensorflow record files containing serialized Tensorflow example protocol buffers \cite{tensorflow2015-whitepaper}. The training set is stored as one collection of Tensorflow record files, and the test set as another. Examples are shuffled randomly (and not necessarily reproducibly) within the Tensorflow record files. Each example is deterministically assigned to either the train or test set using a key feature, such as the conversation thread ID in Reddit, guaranteeing that the same split is created whenever the dataset is generated. By default the train set consists of 90\% of the total data, and the test set the remaining 10\%.

\begin{figure}[htb]
    \centering
    {\small
        \begin{tabular}{>{\bfseries}r >{\itshape}l}
          context/1 & Hello, how are you? \\
          context/0& I am fine. And you? \\
          context & Great. What do you think of the weather? \\
          response & It doesn't feel like February. \\
        \end{tabular}
    }
    
    \caption{
        \label{fig:example}
        An illustrative Tensorflow example in a conversational dataset, consisting of a conversational context and an appropriate response. Each string is stored as a \emph{bytes} feature using its UTF-8 encoding.
    }

\end{figure}

Each Tensorflow example contains a conversational context and a response that goes with that context, see e.g. figure~\ref{fig:example}. Explicitly, each example contains a number of string features:

\begin{itemize}
  \setlength\itemsep{0.1em}
\item A \textbf{context} feature, the most recent text in the conversational context.
\item A \textbf{response} feature, text that is in direct response to the context.
\item A number of extra context features, \textbf{context/0}, \textbf{context/1} etc. going back in time through the conversation. They are named in reverse order so that \textbf{context/\emph{i}} always refers to the $i^\text{th}$ most recent extra context, so that no padding needs to be done, and datasets with different numbers of extra contexts can be mixed.
\item Depending on the dataset, there may be some extra features also included in each example. For instance, in Reddit the author of the context and response are identified using additional features.
\end{itemize}

\section{Datasets}
\label{sec:datasets}

Rather than providing the raw processed data, we provide scripts and instructions to the users to generate the data themselves. This allows for viewing and potentially manipulating the pre-processing and filtering steps. The repository contains instructions for generating datasets with standard parameters split deterministically into train and test portions. These allow for defining reproducible evaluations in research papers. Section~\ref{sec:evaluation} presents benchmark results on these standard datasets for a variety of conversational response selection models.

Dataset creation scripts are written using Apache Beam and Google Cloud Dataflow \cite{Akidau2015}, which parallelizes the work across many machines. Using the default quotas, the Reddit script starts 409 workers to generate the dataset in around 1 hour and 40 minutes. This includes reading the comment data from the BigQuery source, grouping the comments into threads, producing examples from the threads, splitting the examples into train and test, shuffling the examples, and finally writing them to sharded Tensorflow record files.

Table~\ref{tab:datasets} provides an overview of the Reddit, OpenSubtitles and AmazonQA datasets, and figure~\ref{fig:examples} in appendix~\ref{app:examples} gives an illustrative example from each.

\begin{table*}[hbt]
    \centering
    {
        \small
        \begin{tabularx}{0.85 \textwidth}{p{18mm} p{58mm} r r}
            \toprule
             & \textbf{Built from} & \textbf{Training size} & \textbf{Testing size} \\
             \midrule
             {Reddit} &  3.7 billion comments in threaded conversations & 654,396,778 & 72,616,937 \\
             {OpenSubtitles} & over 400 million lines from movie and television subtitles (also available in other languages) &  283,651,561 & 33,240,156 \\
             {AmazonQA} &  over 3.6 million question-response pairs in the context of Amazon products  & 3,316,905 & 373,007 \\
             \bottomrule
        \end{tabularx}
    }
    \caption{
        Summary of the datasets included in the public repository. The Reddit data is taken from January 2015 to December 2018, and the OpenSubtitles data from 2018.
    }
    \label{tab:datasets}
\end{table*}

\subsection{Reddit}


Reddit is an American social news aggregation website, where users can post links, and take part in discussions on these posts. 
Reddit is extremely diverse \cite{Schrading:2015emnlp,AlRfou:2016arxiv}: there are more  than  300,000 sub-forums (i.e., subreddits) covering various topics of discussion. These threaded discussions, available in a public \emph{BigQuery} database,  provide a large corpus of conversational contexts paired with appropriate responses.  Reddit data has been used to create conversational response selection data by \newcite{AlRfou:2016arxiv,Cer:2018arxiv,Yang:2018repl}. We share code that allows generating datasets from the Reddit data in a reproducible manner: with consistent filtering, processing, and train/test splitting. We also generate data using two more years of data than the previous work, 3.7 billion comments rather than 2.1 billion, giving a final dataset with 176 million more examples.

Reddit conversations are threaded. Each post may have multiple top-level comments, and every comment may have multiple children comments written in response. In processing, each Reddit thread is used to generate a set of examples. Each response comment generates an example, where the context is the linear path of comments that the comment is in response to.

Examples may be filtered according to the contents of the context and response features. The example is filtered if either feature has more than 128 characters, or fewer than 9 characters, or if its text is set to \emph{[deleted]} or \emph{[removed]}. Full details of the filtering are available in the code, and configurable through command-line flags.

Further back contexts, from the comment's parent's parent etc., are stored as extra context features. Their texts are trimmed to be at most 128 characters in length, without splitting words apart. This helps to bound the size of an individual example.

The train/test split is deterministic based on the thread ID. As long as all the input to the script is held constant (the input tables, filtering thresholds etc.), the resulting datasets should be identical.

The data from 2015 to 2018 inclusive consists of 3,680,746,776 comments, in 256,095,216 threads. In total, 727,013,715 Tensorflow examples are created from this data.

\subsection{OpenSubtitles}

OpenSubtitles is a growing online collection of subtitles for movies and television shows available in multiple languages.
As a starting point, we use the corpus collected by \newcite{lison2016opensubtitles2016}, originally intended for statistical machine translation. This corpus is regenerated every year, in 62 different languages.

Consecutive lines in the subtitle data are used to create conversational examples. There is no guarantee that different lines correspond to different speakers, or that consecutive lines belong to the same scene, or even the same show. The data nevertheless contains a lot of interesting examples for modelling the mapping from conversational contexts to responses.

Short and long lines are filtered, and some text is filtered such as character names and auditory description text.  The English 2018 data consists of 441,450,449 lines, and generates 316,891,717 examples. The data is split into chunks of 100,000 lines, and each chunk is used either for the train set or the test set.

\subsection{AmazonQA}

This dataset is based on a corpus extracted by \newcite{Wan2016ModelingAS,McAuley2016}, who scraped questions and answers from Amazon product pages. This provides a corpus of question-answer pairs in the e-commerce domain. Some questions may have multiple answers, so one example is generated for each possible answer.

Examples with very short or long questions or responses are filtered from the data, resulting in a total of 3,689,912 examples. The train/test split is computed deterministically using the product ID.

\section{Response Selection Task} \label{sec:methods}

The conversational datasets included in this repository facilitate the training and evaluation of a variety of models for natural language tasks. For instance, the datasets are suitable for training generative models of conversational response \cite{serban2016generative,Ritter:2011,VinyalsL15,SordoniGABJMNGD15,ShangLL15,Kannan:2016kdd}, as well as discriminative methods of conversational response selection \cite{Lowe2015TheUD, inaba2016neural, yu2016strategy,Henderson:2017arxiv}.

\begin{figure*}[t]
    \centering
    \begin{subfigure}[t]{0.48\linewidth}
        \centering
        \includegraphics[width=1.0\linewidth]{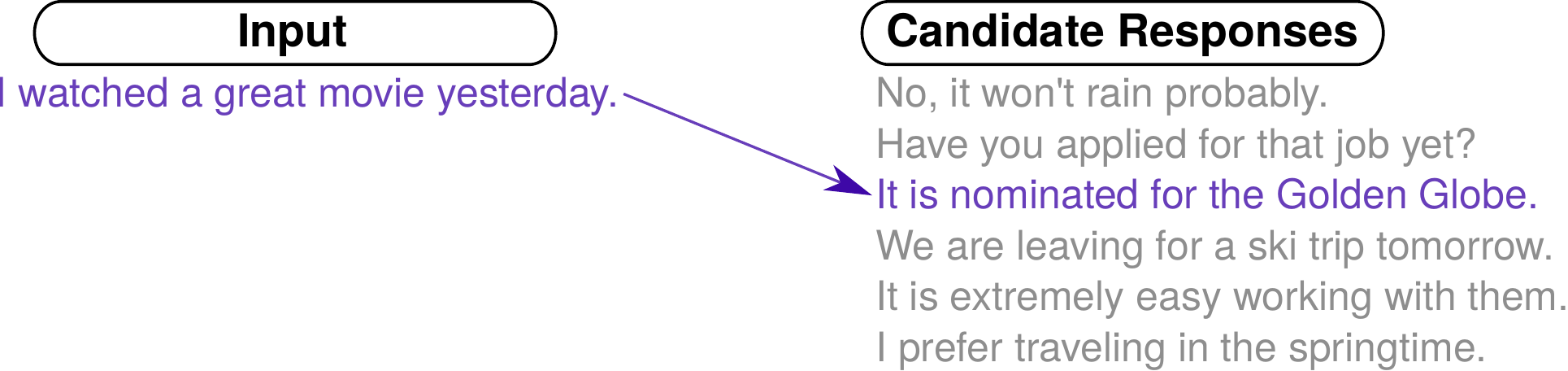}
        \label{fig:ex1}
    \end{subfigure}%
    \hspace{1em}
    \begin{subfigure}[t]{0.48\textwidth}
        \centering
        \includegraphics[width=1.00\linewidth]{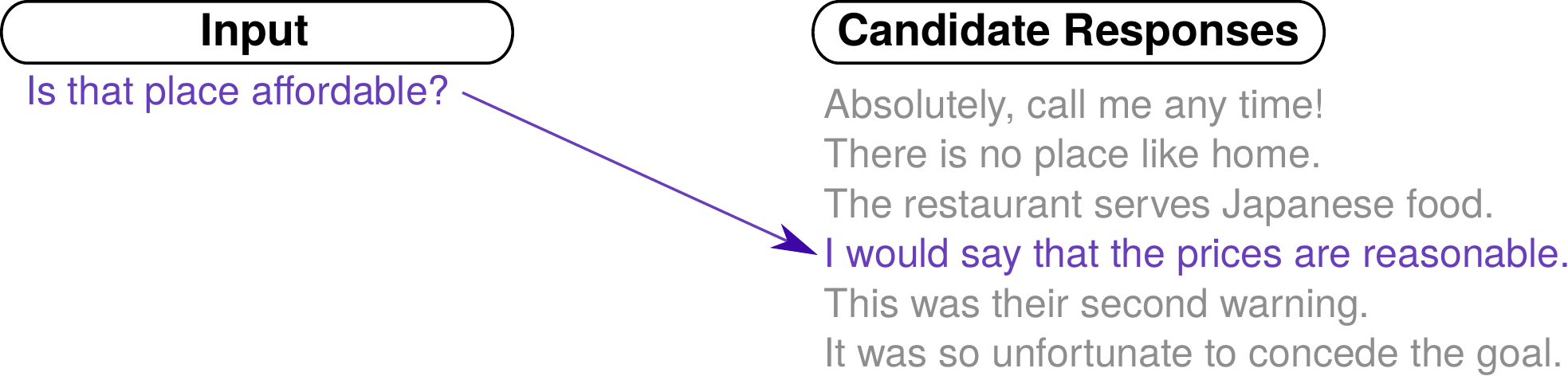}
        \label{fig:ex2}
    \end{subfigure}
    \vspace{-2mm}
	\caption{Two examples illustrating the conversational response selection task: given the input context sentence, the goal is to identify the relevant response from a large pool of candidate responses.}
\vspace{-2.5mm}
\label{fig:selection}
\end{figure*}

The task of conversational response selection is to identify a correct response to a given conversational context from a pool of candidates, as illustrated in figure~\ref{fig:selection}. Such models are typically evaluated using \emph{Recall@k}, a typical metric in information retrieval literature. This measures how often the correct response is identified as one of the top~$k$ ranked responses \cite{Lowe2015TheUD, inaba2016neural, yu2016strategy, AlRfou:2016arxiv, Henderson:2017arxiv, Lowe:2017dd, Wu:2017acl, Cer:2018arxiv, Chaudhuri:2018conll, Du:2018scai, Kumar2018, liu2018customized, Yang:2018repl, Zhou:2018acl, gunasekara2019dstc7, Tao:2019wsdm}.  Models trained to select responses can be used to drive dialogue systems, question-answering systems, and response suggestion systems. The task of response selection provides a powerful signal for learning implicit semantic representations useful for many downstream tasks in natural language understanding \cite{Cer:2018arxiv, Yang:2018repl}.

The \emph{Recall@k} metric allows for direct comparison between models. Direct comparisons are much more difficult for generative models, which are typically evaluated using perplexity scores or using human judgement. Perplexity scores are dependent on normalization, tokenization, and choice of vocabulary, while human judgement is expensive and time consuming.

When evaluating conversational response selection models on these datasets, we propose a \emph{Recall@k} metric termed \emph{1-of-100 accuracy}. This is \emph{Recall@1} using 99 responses sampled from the test dataset as negatives. This \emph{1-of-100 accuracy} metric has been used in previous studies: \cite{AlRfou:2016arxiv, Henderson:2017arxiv, Cer:2018arxiv, Kumar2018, Yang:2018repl, gunasekara2019dstc7}. While there is no guarantee that the 99 randomly selected negatives will all be \emph{bad} responses, the metric nevertheless provides a simple summary of model performance that has been shown to correlate with user-driven quality metrics \cite{Henderson:2017arxiv}. For efficient computation of this metric, batches of 100 (context, response) pairs can be processed such that the other 99 elements in the batch serve as the negative examples.

Sections~\ref{sec:keyword}~and~\ref{sec:vector} present baseline methods of conversational response selection that are implemented in the repository. These baselines are intended to run quickly using a subset of the training data, to give some idea of performance and characteristics of each dataset. Section~\ref{sec:encoder} describes a more competitive neural encoder model that is trained on the entire training set.

\subsection{Keyword-based Methods} \label{sec:keyword}

The keyword-based baselines use keyword similarity metrics to rank responses given a context. These are typical baselines for information retrieval tasks. The \textsc{tf-idf} method computes inverse document frequency statistics on the training set, and scores responses using their tf-idf cosine similarity to the context \cite{Manning:2008ir}.

The \textsc{bm25} method builds on top of the tf-idf similarity, applying an adjustment to the term weights \cite{Robertson:2009}.

\subsection{Vector-based Methods} \label{sec:vector}

The vector-based methods use publicly available neural net embedding models to embed contexts and responses into a vector space. We include the following five embedding models in the evaluation, all of which are available on \href{https://www.tensorflow.org/hub}{Tensorflow Hub}:

\begin{description}[itemindent=1mm]
  \setlength\itemsep{0.1em}
    \item[\textsc{use}] the Universal Sentence Encoder from \newcite{Cer:2018arxiv}
    \item[\textsc{use-large}] a larger version of the Universal Sentence Encoder
    \item[\textsc{elmo}]  the Embeddings from Language Models approach from \newcite{Peters:2018}.
    \item[\textsc{bert-small}] the deep bidirectional transformer model of \newcite{Devlin:2018arxiv}.
    \item[\textsc{bert-large}] a larger deep bidirectional transformer model.
\end{description}

There are two vector-based baseline methods, one for each of the above models. The \textsc{sim} method ranks responses according to their cosine similarity with the context vector. This method relies on pretrained models and does not use the training set at all.

The \textsc{map} method learns a linear mapping on top of the response vector. The final score of a response with vector $\mathbf{y}$ given a context with vector $\mathbf{x}$ is the cosine similarity $\langle \cdot,\: \cdot\rangle$ of the context vector with the mapped response vector:
\begin{align}
\langle \mathbf{x},\: \left( W + \alpha I \right) \cdot\mathbf{y} \rangle
\end{align}
where $W, \: \alpha$ are learned parameters and $I$ is the identity matrix. This allows learning an arbitrary linear mapping on the context side, while the residual connection gated by $\alpha$ makes it easy for the model to interpolate with the \textsc{SIM} baseline. Vectors are L2-normalized before being fed to the \textsc{map} method, so that the method is invariant to scaling.

The $W$ and $\alpha$ parameters are learned on a random sample of 10,000 examples from the training set, using the dot product loss from \newcite{Henderson:2017arxiv}. A sweep over learning rate and regularization parameters is performed using a held-out development set. The final learned parameters are used on the evaluation set.

The combination of the three embedding models with the two vector-based methods results in the following six baseline methods: \textsc{use-sim, use-map, use-large-sim, use-large-map, elmo-sim,} and \textsc{elmo-map}.

\subsection{Encoder Model} \label{sec:encoder}

We also train and evaluate a neural encoder model that maps the context and response through separate sub-networks to a shared vector space, where the final score is a dot-product between a vector representing the context and a vector representing the response as per \newcite{Henderson:2017arxiv, Cer:2018arxiv, Kumar2018, Yang:2018repl}. This model is referred to as \textsc{\encoder} in the evaluation.

Full details of the neural structure are given in \newcite{Henderson2019b}. To summarize, the context and response are both separately passed through sub-networks that:

\begin{enumerate}
  \setlength\itemsep{0.1em}
    \item split the text into unigram and bigram features
    \item convert unigrams and bigrams to numeric IDs using a vocabulary of known features in conjunction with a hashing strategy for unseen features
    \item separately embed the unigrams and bigrams using large embedding matrices
    \item separately apply self-attention then reduction over the sequence dimension to the unigram and bigram embeddings
    \item combine the unigram and bigram representations, then pass them through several dense hidden layers
    \item L2-normalize the final hidden layer to obtain the final vector representation
\end{enumerate}

Both sub-networks are trained jointly using the dot-product loss of \newcite{Henderson:2017arxiv}, with label smoothing and a learned scaling factor.

\section{Evaluation} \label{sec:evaluation}

All the methods discussed in section~\ref{sec:methods} are evaluated on the three standard datasets from section~\ref{sec:datasets}, and the results are presented in table~\ref{tab:evaluation}. In this evaluation, all methods use only the (immediate) context feature to score the responses, and do not use other features such as the extra contexts.

\begin{table*}[htb]

\begingroup
\renewcommand*{\arraystretch}{1.2}

\centering
\newcolumntype{Y}{>{\centering\arraybackslash}X}
\begin{tabularx}{0.85 \linewidth}{l YYY}
\toprule
{} & {{Reddit}} & {{OpenSubtitles}} &  {{AmazonQA}}  \\ \midrule
{\textsc{tf-idf}} &          {26.7} & {10.9} & {51.8}  \\
{\textsc{bm25}} &            {27.6} & {10.9} & {52.3}  \\
\hdashline

\addlinespace[1ex] {\textsc{use-sim}} & {36.6} & {13.6} & {47.6} \\
{\textsc{use-map}} &         {40.8} & {15.8} & {54.4} \\
{\textsc{use-large-sim}} &   {41.4} & {14.9} & {51.3} \\
{\textsc{use-large-map}} &   {47.7} & {18.0} & {61.9} \\
{\textsc{elmo-sim}} &        {12.5} & \makebox[\widthof{99.9}][r]{9.5}  & {16.0} \\
{\textsc{elmo-map}} &        {19.3} & {12.3} & {33.0} \\
{\textsc{bert-small-sim}} &  {17.1}    & {13.8} & {27.8} \\
{\textsc{bert-small-map}} & {24.5}     & {17.5}& {45.8}\\
{\textsc{bert-large-sim}} & {14.8}     & {12.2}& {25.9}\\
{\textsc{bert-large-map}} & {24.0}     & {16.8}&{44.1} \\

\hdashline

\addlinespace[1ex] {\textsc{\encoder}}    & \textbf{61.3} & \textbf{30.6} & \textbf{84.2}  \\
\bottomrule
\end{tabularx}
\vspace{1em}
\caption{
    \emph{1-of-100 accuracy} results for keyword-based baselines, vector-based baselines, and the encoder model for each of the three standard datasets. The latest evaluation results are maintained in the \href{https://github.com/PolyAI-LDN/conversational-datasets/blob/master/BENCHMARKS.md}{repository}. Results are computed on a random subset of 50,000 examples from the test set (500 batches of 100).
}
\label{tab:evaluation}
\endgroup
\end{table*}

The keyword-based \textsc{tf-idf} and \textsc{bm25} are broadly competitive with the vector-based methods, and are particularly strong for AmazonQA, possibly because rare words such as the product name are informative in this domain. Learning a mapping with the \textsc{map} method gives a consistent boost in performance over the \textsc{sim} method, showing the importance of learning the mapping from context to response versus simply relying on similarity. This approach would benefit from more data and a more powerful mapping network, but we have constrained the baselines so that they run quickly on a single computer. The Universal Sentence Encoder model outperforms ELMo in all cases.

The \textsc{\encoder} model significantly outperforms all of the baseline methods. This is not surprising, as it is trained on the entire training set using multiple GPUs for several hours. We welcome other research groups to share their results, and we will be growing the table of results in the repository.

\section{Conclusion}

This paper has introduced a repository of conversational datasets, providing hundreds of millions examples for training and evaluating conversational response selection systems under a standard evaluation framework. Future work will involve introducing more datasets in this format, more competitive baselines, and more benchmark results. We welcome contributions from other research groups in all of these directions.

\bibliographystyle{acl_natbib}  
\bibliography{references}

\cleardoublepage
\onecolumn
\appendix

\section{Appendix} \label{app:examples}

\begin{figure}[htb]
    \centering
    {
        \small
        \begin{tabular}{>{\bfseries}r >{\itshape}p{100mm}}
          \multicolumn{2}{l}{\textbf{Reddit}} \\
          context/2 & Could someone there post a summary of the insightful moments. \\
          context/1 & Basically L2L is the new deep learning. \\
          context/0 & What's ``L2L'' mean? \\
          context & ``Learning to learn'', using deep learning to design the architecture of another deep network: https://arxiv.org/abs/1606.04474 \\
          response & using deep learning with SGD to design the learning algorithms of another deep network   * \\[1ex]
          context\_author & goodside \\
          response\_author & NetOrBrain \\
          subreddit & MachineLearning \\
          thread\_id$\star$ & \href{https://www.reddit.com/r/MachineLearning/comments/5h6yvl/d_recurrent_neural_networks_and_other_machines/daz60rv/?context=8&depth=9}{5h6yvl} \\
          \midrule
          
          \multicolumn{2}{l}{\textbf{OpenSubtitles}} \\
          context/9 & So what are we waiting for? \\
          context/8 & Nothing, it... \\
          context/7 & It's just if... \\
          context/6 & If we've underestimated the size of the artifact's data stream... \\
          context/5 & We'll fry the ship's CPU and we'll all spend the rest of our lives stranded in the Temporal Zone. \\
          context/4 & The ship's CPU has a name. \\
          context/3 & Sorry, Gideon. \\
          context/2 & Can we at least talk about this before you connect... \\
          context/1 & Gideon? \\
          context/0 & You still there? \\
          context & Oh my God, we killed her. \\
          response & Artificial intelligences cannot, by definition, be killed, Dr. Palmer. \\[1ex]
          file\_id$\star$ & lines-emk \\
          \midrule

          \multicolumn{2}{l}{\textbf{AmazonQA}} \\
          context & i live in singapore so i would like to know what is the plug cos we use those 3 pin type \\
          response & it's a 2 pin U.S. plug, but you can probably get an adapter , very good hair dryer! \\[1ex]
          product\_id$\star$ & \href{https://www.amazon.com/ask/questions/asin/B003XNYHWS/1/ref=ask_ql_psf_ql_hza?sort=SUBMIT_DATE&isAnswered=true}{B003XNYHWS} \\
          
        \end{tabular}
        \caption{Examples from the three datasets. Each example is a mapping from feature names to string features. Features with a star $\star$ are used to compute the deterministic train/test split.}
        \label{fig:examples}
    }
\end{figure}

\end{document}